%% file: neurips_2026.tex
\documentclass{article}

 \usepackage[preprint]{neurips_2026}

\usepackage[utf8]{inputenc} 
\usepackage[T1]{fontenc}    
\usepackage{hyperref}       
\usepackage{url}            
\usepackage{booktabs}       
\usepackage{amsfonts}       
\usepackage{nicefrac}       
\usepackage{microtype}      
\usepackage{xcolor}         
\usepackage{amsmath}
\usepackage[pdftex]{graphicx}
\usepackage{cleveref}
\usepackage{multirow}
\usepackage{booktabs}
\usepackage{subcaption}
\usepackage{wrapfig}
\usepackage{tikz}
\usepackage{tabularx}
\usepackage{placeins}

\definecolor{Gray}{gray}{0.9}
\definecolor{LightCyan}{rgb}{0.88,1,1}
\definecolor{red}{RGB}{144,0,32}
\definecolor{blue}{rgb}{0.06, 0.3, 0.57}
\definecolor{warmblack}{rgb}{0.0, 0.26, 0.26}
\definecolor{purple}{rgb}{0.4, 0.01, 0.24}
\definecolor{tawny}{rgb}{0.8, 0.34, 0.0}

\hypersetup{
    colorlinks=true,
    citecolor=tawny,
    linkcolor=purple,
    urlcolor=purple,
    }

\title{Procedural Pretraining for Molecular Property Prediction}

\author{%
  Moritz Friedemann$^{1}$ \quad
  Zachary Shinnick$^{1,2}$ \quad
  Philip Torr$^{1,3}$ \quad
  Bruno Andreis$^{1,3}$ \\[0.6em]
  $^{1}$University of Oxford \quad
  $^{2}$Adelaide University \quad
  $^{3}$Slater Labs
}

\begin{document}

\maketitle

\input{abstract}
\input{introduction}

\input{related}

\input{method}

\input{experiments}
\input{discussion}

\newpage
\begingroup
\small
\bibliographystyle{plainnat}
\bibliography{references}
\endgroup

\clearpage
\input{appendix}

\end{document}

%% file: abstract.tex
\begin{abstract}
Molecular property prediction is often limited by the small size of labeled downstream datasets, motivating pretraining on large corpora of unlabeled molecules. In this work, we ask whether useful inductive biases can instead be learned from abstract, procedurally generated data before a model sees any molecular data. We introduce a three-stage training pipeline consisting of procedural pretraining, molecular pretraining on SMILES, and downstream fine-tuning, and evaluate several procedural tasks spanning sequence structure, cellular automata, and graph reasoning. We find that procedural pretraining can improve molecular property prediction even after subsequent molecular pretraining: on Lipophilicity, \textsc{Reverse} reduces test error by 4.8\%. For context, the magnitude of this improvement is roughly 90\% of the performance difference between our 250K-molecule baseline and the publicly released MoLFormer checkpoint pretrained on approximately 100M molecules. Our analysis shows that the benefit is strongest under downstream data scarcity, depends on the structure of the procedural data rather than only surface-level statistics, and does not increase monotonically with additional procedural training. Instead, transfer typically peaks at an intermediate procedural budget and deteriorates as the model approaches convergence on the procedural task. We further find that, for several tasks, much of the transferable information is localized in the attention layers, while feed-forward layers can contribute to over-specialization. These results show that procedural data can provide transferable structure for molecular learning and offer a complementary route to improving performance when labeled molecular data are limited.

\end{abstract}

%% file: introduction.tex
\section{Introduction}

\begin{wrapfigure}{r}{0.52\columnwidth}
    \centering
    \vspace{-20pt}
    \includegraphics[width=\linewidth]{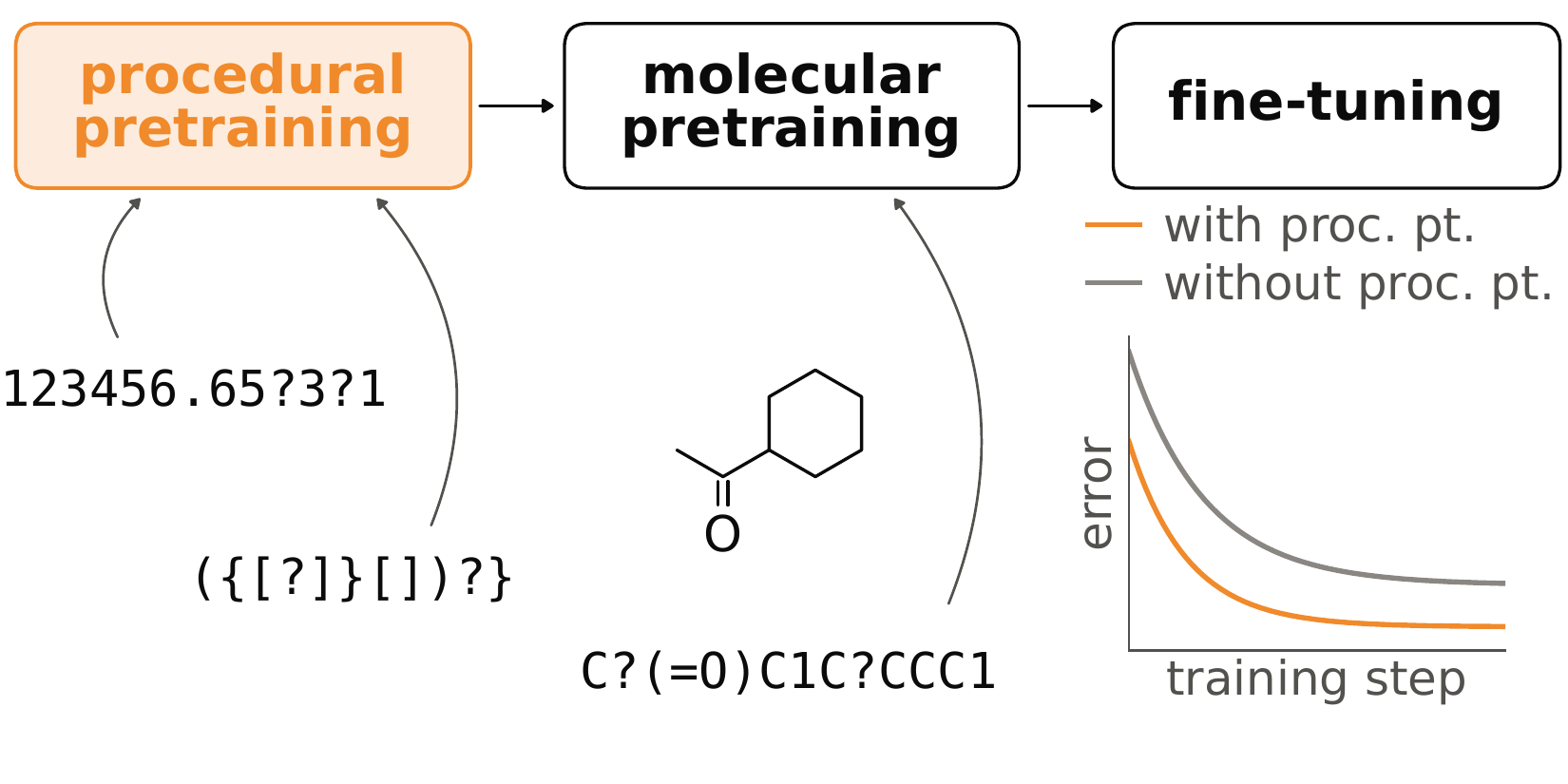}
    \label{fig:selective-transfer}
    \vspace{-16pt}
    \caption{Overview of the proposed three-stage training pipeline. A transformer is first pretrained on procedurally generated masked-token tasks, then pretrained on molecular SMILES, and finally fine-tuned for molecular property prediction. Procedural pretraining can reduce downstream error relative to molecular pretraining and fine-tuning alone.}
    \vspace{-15pt}
\end{wrapfigure}

Modern molecular machine learning models are often trained in a regime where unlabeled molecules are abundant \citep{irwin_zinc_2004,kim_pubchem_2019} but labeled downstream data are scarce. Molecular property prediction datasets frequently contain only hundreds or thousands of labeled examples \citep{wu_moleculenet_2018}, making it difficult for high-capacity models to learn useful representations from downstream supervision alone. A common response has been to adopt the pretraining paradigm developed for language models: models are first trained on large collections of unlabeled molecular representations, such as SMILES strings, and subsequently fine-tuned for a target property \citep{chithrananda_chemberta_2020,ross_large-scale_2022}. While such molecular pretraining can substantially improve performance, acquiring and processing increasingly large molecular corpora can be expensive, and it remains unclear whether useful inductive biases must themselves be learned from molecular data.

Recent work suggests an alternative source of pretraining signal: procedurally generated data. Simple algorithmic processes, such as logical operations~\citep{wu_insights_2022,shinnick2025transformers}, formal languages~\citep{papadimitriou2023injecting,hu_between_2025}, and cellular automata~\citep{zhang2025intelligence,lee_training_2026}, can generate effectively unlimited data with controlled structural properties. Procedural pretraining on such data has been shown to induce representations that transfer to natural language and vision tasks~\citep{jiang_procedural_2026,shinnick_can_2026}. This raises a surprising possibility for molecular learning: \textit{can a model learn useful computational structure before seeing a single molecule, and can that structure subsequently improve molecular property prediction?}

We investigate this question by introducing a procedural pretraining stage before conventional molecular pretraining and downstream fine-tuning. Using MoLFormer~\citep{ross_large-scale_2022} as our base architecture, we pretrain on several procedural tasks spanning sequence manipulation \citep{jiang_procedural_2026}, hierarchical structure \citep{hu_between_2025}, dynamical systems \citep{lee_training_2026}, and graph reasoning, then transfer the learned transformer weights to molecular prediction. Our experiments consider both a realistic three-stage pipeline, procedural pretraining, molecular pretraining, and downstream fine-tuning, and a simplified setting that removes molecular pretraining to isolate the mechanisms responsible for transfer.

We find that procedural pretraining can improve molecular property prediction even after subsequent pretraining on real molecular data. On Lipophilicity \citep{wu_moleculenet_2018}, pretraining on a simple sequence-reversal task reduces test error by 4.8\% relative to a molecular-pretraining-only baseline, recovering roughly 90\% of the performance difference between our 250K-molecule baseline and the publicly released MoLFormer checkpoint pretrained on approximately 100M molecules. We then use controlled experiments to characterize when this transfer is beneficial, what properties of the procedural data drive it, and where the transferable information is represented in the model.

Our main contributions are:

\begin{itemize}
    \item We show that \textbf{procedural pretraining can improve molecular property prediction even after subsequent molecular pretraining}, demonstrating that useful non-molecular structure can survive a realistic domain-specific pretraining stage.

    \item We find that \textbf{structural alignment} between procedural and molecular data provides an initial head start during molecular pretraining but \textbf{does not reliably predict downstream benefit}.

    \item We show that the benefit is \textbf{largest under downstream data scarcity}, suggesting that procedural pretraining acts as an inductive bias that is most valuable when labeled molecular data are limited.

    \item We find that \textbf{more procedural pretraining is not always better}. Across four of five tasks, downstream performance peaks at an intermediate procedural budget and deteriorates as the model approaches the optimum of the procedural task.

    \item We demonstrate that transfer depends on the \textbf{structure of the procedural data rather than surface-level statistics}: shuffling token order while preserving vocabulary, sequence length, and token frequencies removes the benefit.

    \item We \textbf{localize much of the transferable information to the attention layers} for several procedural tasks, while finding that feed-forward layers can contribute to the over-specialization observed at larger procedural budgets.
\end{itemize}

%% file: related.tex
\section{Related Work}

\paragraph{Pretraining on procedural data.}
Recent work has shown that useful representations can be learned from procedurally generated data rather than natural corpora. In language, pretraining on artificial languages, formal languages, neural cellular automata, and simple algorithmic tasks can improve downstream performance on natural language tasks \citep{papadimitriou_learning_2020,hu_between_2025,lee_training_2026,jiang_procedural_2026}. Similar results have been observed in vision using synthetic images such as fractals or structured noise \citep{kataoka_pre-training_2021,baradad_jurjo_learning_2021}. While many of these procedures reproduce structural properties of the target domain, recent work demonstrates transfer even across domains: sequence-based formal languages can improve vision transformers \citep{shinnick_can_2026}, and music pretraining can benefit language models \citep{nomura_listen_2026}. This suggests that procedural pretraining can provide useful inductive biases without closely resembling downstream data.

Prior work has also investigated which properties make procedural pretraining transferable. Structural properties such as diversity, long-range dependencies, and hierarchical organization can improve transfer \citep{baradad_jurjo_learning_2021,chiang_transferability_2021,hu_between_2025}. However, strong domain alignment is not always necessary: even simple sequence operations retain part of the benefit of natural-language pretraining \citep{wu_insights_2022}, and transferable information may reside in different transformer components depending on the downstream task \citep{jiang_procedural_2026}. Together, these findings point to two possible sources of transfer: alignment between the procedural and target distributions, and more general changes to the model's inductive bias.

We extend this line of work to molecular modeling, where labeled data are comparatively scarce and procedural pretraining has received little attention. We study whether abstract procedural tasks can complement molecular pretraining, how alignment with SMILES structure affects transfer, and which model components and properties of procedural-task learning account for downstream gains.

%% file: method.tex
\section{Methodology}\label{sec:methodology}
\paragraph{Training pipeline.}
We study a three-stage training pipeline consisting of
(a) procedural pretraining,
(b) molecular pretraining, and
(c) downstream fine-tuning.
The procedural stage exposes the model only to algorithmically generated data, before it encounters any molecular examples.
We then optionally continue pretraining on unlabeled SMILES strings before fine-tuning on a molecular property prediction task.
In \Cref{res:when-and-why}, we omit the molecular pretraining stage for some experiments and fine-tune directly after procedural pretraining in order to isolate the effect of the procedural data.

\paragraph{Procedural pretraining tasks.}
We consider five procedural tasks designed to expose the model to different forms of structure.
Three operate directly on synthetic sequences: \textsc{Reverse}, in which the model must recover the reversed form of a token sequence; \textsc{Dyck}, which consists of properly nested bracket sequences; and \textsc{Dyck-Shuffle}, which permits crossing bracket dependencies. \textsc{Dyck} mirrors the structure of branching parentheses in SMILES, while \textsc{Dyck-Shuffle} mirrors ring-closure digits.
We additionally consider neural cellular automaton (\textsc{NCA}) trajectories, which introduce spatiotemporal structure, and a shortest-path (\textsc{SSP}) task in which the model predicts node distances from the SMILES representation of a random graph.
All tasks are formulated as masked-token prediction problems so that the same model architecture and training objective can be used throughout procedural and molecular pretraining.
Examples of the procedural inputs are shown in \Cref{fig:procedural-tasks}.

\input{procedural-task-examples}

\paragraph{Model.}
We use MoLFormer \citep{ross_large-scale_2022}, an encoder-only transformer with approximately $44$M parameters designed for learning from SMILES representations.
MoLFormer uses rotary positional embeddings (RoPE), linear attention based on generalized random features, and a regular-expression-based SMILES tokenizer with a vocabulary of 2,362 tokens.
We use the same transformer architecture across all training stages.

\paragraph{Molecular pretraining.}
MoLFormer is conventionally pretrained using masked token modeling (MTM) on large collections of unlabeled SMILES strings drawn from PubChem \citep{kim_pubchem_2019} and ZINC \citep{irwin_zinc_2004}.
The publicly released MoLFormer checkpoint was trained on approximately $100$M molecules, which is beyond our available compute budget.
To evaluate procedural pretraining within a realistic molecular pretraining pipeline, we therefore construct a smaller-scale approximation of the setup in \citet{ross_large-scale_2022}.
We pretrain for 12 epochs on a 250k-molecule subset of PubChem using the same MTM objective and a masking probability of $0.15$.
When comparing procedural and non-procedural initializations, all models receive the same molecular pretraining data and compute budget.

\paragraph{Downstream evaluation.}
We evaluate on three molecular property prediction benchmarks with continuous regression targets:
Lipophilicity from MoleculeNet \citep{wu_moleculenet_2018}, containing approximately $4$k molecules;
FreeSolv, containing 642 molecules;
and QM9, for which we predict the HOMO--LUMO gap and use either the full dataset or controlled subsets of the training set.
Models are fine-tuned for 500 epochs.
We report test mean absolute error (MAE) on standardized targets from the checkpoint achieving the best validation performance, following \citet{ross_large-scale_2022}.
Unless stated otherwise, comparisons use identical downstream splits and optimization settings.

\paragraph{Weight transfer.}
We partition the model parameters into token embeddings, attention projections, feed-forward layers, LayerNorm parameters, and the output head.
The procedural and molecular domains use different token semantics, so we do not transfer learned procedural token embeddings.
Instead, the embedding matrix is held fixed at its random initialization during procedural pretraining, ensuring that any transferable information must be encoded in the transformer backbone rather than in task-specific token representations.
We also discard the procedural output head, whose dimensionality and semantics need not match those of subsequent stages.
Our default transfer setting therefore retains the learned transformer backbone while reinitializing the input and output interfaces as required.

To localize where transferable information is stored, we additionally consider selective transfer.
We transfer either only the attention-layer weights or only the feed-forward-layer weights, while reinitializing all remaining parameters, including LayerNorms.
This allows us to measure how much of the downstream benefit can be attributed to each component of the transformer.

%% file: procedural-task-examples.tex
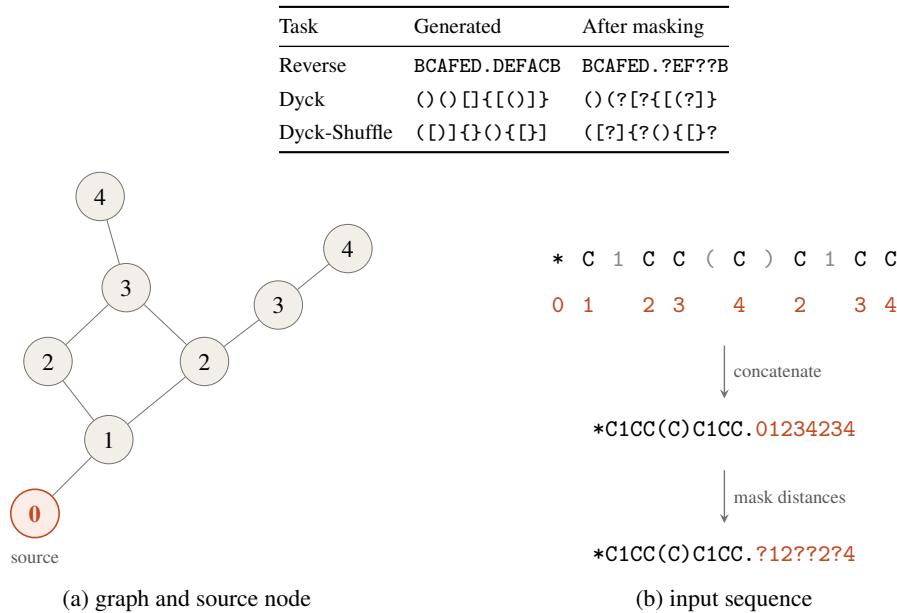
\begin{figure}[t]
    \centering
    \begin{minipage}{0.6\linewidth}
        \centering
        \footnotesize
        \setlength{\tabcolsep}{4pt}
        \begin{tabular}{@{}l l l@{}}
        \toprule
        Task & Generated & After masking \\
        \midrule
        Reverse      & \ttfamily BCAFED.DEFACB   & \ttfamily BCAFED.?EF??B \\[3pt]
        Dyck         & \ttfamily ()()[]\{[()]\}  & \ttfamily ()(?[?\{[(?]\} \\[3pt]
        Dyck-Shuffle & \ttfamily ([)]\{\}()\{[\}] & \ttfamily ([?]\{?()\{[\}? \\
        \bottomrule
        \end{tabular}
    \end{minipage}

    \vspace{6pt}

    \input{ssp_task_fig}
    \caption{Procedural pretraining tasks.
    Top: example sequences of the sequence-level tasks before and after masking.
    Bottom: how an input sequence for the \textsc{SSP} task is created. a) A molecule-like random graph is generated and a source node chosen at random. b) The SMILES representation of the graph (using \texttt{*} for the source and \texttt{C} for all other nodes) is concatenated with shortest-path distances from the source to all other nodes. }
    \label{fig:procedural-tasks}
\end{figure}

%% file: ssp_task_fig.tex
\definecolor{distcol}{HTML}{C0491F}   
\definecolor{srcfill}{HTML}{FAECE7}
\definecolor{nodefill}{HTML}{F1EFE8}
\definecolor{tokgray}{HTML}{8A8880}   
\definecolor{edgegray}{HTML}{73726C}

\begin{subfigure}[b]{0.40\textwidth}
\centering
\begin{tikzpicture}[x=1.15cm,y=1.15cm,
  atom/.style={circle,draw=edgegray,line width=0.35pt,fill=nodefill,
               minimum width=6.4mm,minimum height=6.4mm,inner sep=0pt,
               font=\small},
  source/.style={atom,draw=distcol,line width=0.6pt,fill=srcfill,
                 text=distcol,font=\small\bfseries},
  bond/.style={draw=edgegray,line width=0.35pt}
]
  \node[source] (a1) at (0.00,0.00) {0};
  \node[atom]   (a2) at (0.85,0.85) {1};
  \node[atom]   (a3) at (0.15,1.75) {2};
  \node[atom]   (a4) at (1.05,2.60) {3};
  \node[atom]   (a5) at (0.75,3.65) {4};
  \node[atom]   (a6) at (1.95,1.75) {2};
  \node[atom]   (a7) at (2.80,2.40) {3};
  \node[atom]   (a8) at (3.60,3.05) {4};

  \draw[bond] (a1)--(a2); \draw[bond] (a2)--(a3);
  \draw[bond] (a3)--(a4); \draw[bond] (a4)--(a5);
  \draw[bond] (a4)--(a6); \draw[bond] (a6)--(a2);
  \draw[bond] (a6)--(a7); \draw[bond] (a7)--(a8);

  \node[font=\scriptsize,text=black!65,anchor=north] at (0,-0.36) {source};
\end{tikzpicture}
\caption{graph and source node}
\end{subfigure}
\hfill
\begin{subfigure}[b]{0.57\textwidth}
\centering
\begin{tikzpicture}[
  x=0.40cm,
  tok/.style={font=\small\ttfamily,anchor=base},
  str/.style={tok,text=tokgray},
  dist/.style={font=\small\ttfamily\bfseries,text=distcol,anchor=base},
  flow/.style={->,>=stealth,draw=edgegray,line width=0.5pt},
  lbl/.style={font=\scriptsize,text=black!65,anchor=west,inner sep=0pt}
]
  
  \foreach \i/\c in {0/*,1/C,3/C,4/C,6/C,8/C,10/C,11/C}
    \node[tok] at (\i,0) {\c};
  \foreach \i/\c in {2/1,5/(,7/),9/1}
    \node[str] at (\i,0) {\c};

  \foreach \i/\d in {0/0,1/1,3/2,4/3,6/4,8/2,10/3,11/4}
    \node[dist] at (\i,-0.60) {\d};

  \draw[flow] (5.5,-1.05) -- (5.5,-1.75);
  \node[lbl]  at (5.8,-1.40) {concatenate};
  \node[font=\small\ttfamily,anchor=base] at (5.5,-2.25)
    {*C1CC(C)C1CC.\textcolor{distcol}{\textbf{01234234}}};

  \draw[flow] (5.5,-2.70) -- (5.5,-3.40);
  \node[lbl]  at (5.8,-3.05) {mask distances};
  \node[font=\small\ttfamily,anchor=base] at (5.5,-3.90)
    {*C1CC(C)C1CC.\textcolor{distcol}{\textbf{?12??2?4}}};
\end{tikzpicture}
\caption{input sequence}
\end{subfigure}

%% file: experiments.tex
\section{Experiments}
\subsection{\label{results:ppt-transfers}Procedural Pretraining Improves Molecular Property Prediction}

We first evaluate procedural pretraining in the full three-stage pipeline: procedural pretraining, molecular pretraining, and downstream fine-tuning. We compare models initialized from each procedural task against a baseline trained only on molecular data, while keeping the molecular pretraining and downstream fine-tuning stages fixed. We evaluate on Lipophilicity, FreeSolv, and a 1k-molecule subset of QM9 for HOMO--LUMO gap prediction.

\input{merged-downstream-table}

\Cref{tab:main_results} reports standardized test MAE and the relative change with respect to the molecular-pretraining-only baseline. On Lipophilicity, two procedural tasks yield clear improvements: \textsc{Reverse} reduces MAE by $4.80\%$ and \textsc{Dyck-Shuffle} by $3.53\%$, with both improvements separated from the baseline across seeds. The remaining three tasks also improve the mean performance, although their gains lie within seed variation. \textsc{Reverse} additionally improves performance on the 1k-molecule QM9 subset by $1.91\%$.

The magnitude of the Lipophilicity improvement is notable relative to scaling molecular pretraining itself. Fine-tuning the publicly released MoLFormer checkpoint, pretrained on approximately $100$M molecules, improves over our 250k-molecule molecular-pretraining baseline by $5.35\%$. By comparison, only 1.25k procedural pretraining steps on \textsc{Reverse} recover a $4.80\%$ improvement without using any additional molecular data.

We do not observe a clear benefit on FreeSolv. Four procedural tasks perform worse than the molecular-pretraining-only baseline, although the released MoLFormer checkpoint, trained on approximately $400\times$ more molecular data, is also worse by $12.01\%$. Given that FreeSolv contains only 642 molecules, the dataset appears too small to reliably resolve differences of this magnitude.

Overall, these results show that information learned from procedural data can survive subsequent molecular pretraining and improve downstream molecular property prediction.

\subsection{\label{res:when-and-why}When and Why Does Procedural Pretraining Help?}

Having established that procedural pretraining can improve the full three-stage molecular learning pipeline, we next investigate when and why this transfer occurs. Unless otherwise noted, experiments in this section use a two-stage diagnostic setting: procedural pretraining followed directly by downstream fine-tuning, with no intervening molecular pretraining. In this setting, procedurally pretrained models are compared against an identically fine-tuned, randomly initialized baseline. Removing the molecular stage isolates the effects of procedural pretraining from representations subsequently learned from SMILES data. Two subsections also examine the full pipeline: \Cref{res:structural-alignment} retains molecular pretraining throughout, while \Cref{exp:requires-structure} reports shuffled-token controls in both settings.

\subsubsection{\label{res:structural-alignment}Does Structural Alignment With Molecules Matter?}

\input{pretraining_dynamics_scarcity}

\textsc{Reverse} provides the strongest downstream improvement in \Cref{results:ppt-transfers}, despite having no obvious structural correspondence with SMILES. By contrast, \textsc{Dyck}, \textsc{Dyck-Shuffle}, and \textsc{SSP} were chosen to reflect structural properties that are more closely aligned with molecular representations. We therefore ask whether such alignment provides an advantage during molecular pretraining.

We compare held-out masked-token modeling loss throughout 12 epochs of molecular pretraining on PubChem250k. As shown in \Cref{fig:molecular-pretraining-loss}, the SMILES-aligned tasks, \textsc{Dyck}, \textsc{Dyck-Shuffle}, and \textsc{SSP}, provide the largest advantage at the beginning of molecular pretraining. However, this advantage decays rapidly, and after several epochs their performance becomes similar to that of \textsc{Reverse}. \textsc{SSP} eventually becomes detrimental despite operating on real SMILES strings.


These results suggest that structural alignment provides an initial head start during molecular pretraining, but that much of this advantage can be recovered from SMILES data alone. More broadly, structural alignment is not a reliable predictor of which tasks will transfer best: \textsc{Reverse} provides the strongest downstream improvement after the full pipeline despite having little obvious alignment with SMILES. Thus, within our task suite, procedural tasks do not need to be structurally aligned with the molecular target domain to transfer effectively. 

\subsubsection{\label{res:data-scarcity}The Benefit Is Largest Under Data Scarcity}

If procedural pretraining supplies useful structure that would otherwise have to be inferred from downstream molecular data, its value should increase as the amount of labeled data decreases. To test this hypothesis, we subsample the QM9 training set to 1k, 16k, 64k, and the full 108,446 molecules. At each size, we compare a model pretrained on \textsc{Dyck} against a randomly initialized baseline and evaluate both models on the same held-out test set of 13,389 molecules.

\Cref{fig:data-scarcity} shows a clear relationship between downstream dataset size and transfer benefit: the advantage of procedural pretraining grows monotonically as the amount of training data decreases. Procedural pretraining therefore acts most strongly as an inductive bias in low-data regimes, where the downstream dataset provides less opportunity for the model to learn the relevant structure directly.

\subsubsection{\label{res:overtraining}More Procedural Pretraining Is Not Always Better}

\input{downstream_pp_budgets}

A practical appeal of procedural data is that it can be generated in effectively unlimited quantities. We therefore ask whether increasing the procedural pretraining budget continually improves downstream transfer.

We vary both task complexity and procedural training duration. For \textsc{Dyck} and \textsc{Dyck-Shuffle}, complexity is controlled by the number of distinct bracket types; for \textsc{Reverse}, by sequence length; and for \textsc{NCA}, by the gzip complexity of the generated trajectories. For each combination of task complexity and training budget, we procedurally pretrain a model and compare its downstream Lipophilicity performance against the randomly initialized baseline.

As shown in \Cref{fig:overtraining}, four of the five procedural tasks exhibit the same qualitative pattern: downstream benefit initially increases with additional procedural training, reaches a maximum at an intermediate budget, and then deteriorates. At the largest budgets, performance is often worse than the randomly initialized baseline. \textsc{NCA} is the only exception and is also the only task that the model does not learn to solve during procedural pretraining.

This behavior suggests that transfer degrades as the model becomes increasingly specialized to the procedural objective. To examine this more directly, we compare downstream benefit with the procedural loss reached at each training budget. For deterministic tasks such as \textsc{Dyck}, \textsc{Reverse}, and \textsc{SSP}, the Bayes-optimal loss is zero, and the models approach this value at large training budgets. \textsc{Dyck-Shuffle} is not perfectly predictable because closing-bracket identity contains aleatoric uncertainty: at a masked position, multiple currently open bracket types can yield a valid sequence. Its Bayes-optimal loss is therefore non-zero and can be computed from the generating process. For $k=128$, this value is approximately $1.267$, which the model approaches at the largest budgets. Procedural losses and Bayes-optima can be found in Appendix~\ref{app:bayes-floor}.

The relevant factor is therefore not low procedural loss in absolute terms, but proximity to the best achievable solution for the procedural task. Across tasks, downstream transfer is strongest before the model approaches procedural convergence, suggesting that useful general structure is acquired earlier than task-specific specialization.

\subsubsection{\label{exp:requires-structure}Transfer Requires Procedural Structure}

One alternative explanation for the observed gains is that procedural pretraining need not teach meaningful structure at all. Any additional pretraining could move the model away from its random initialization, calibrate parameter scales, or expose it to relevant sequence lengths and token-frequency statistics. To distinguish these effects from learning structured dependencies, we construct shuffled-token controls. We randomly permute token order within each procedural sequence, thereby destroying order-dependent structure while preserving the vocabulary, sequence lengths, and token frequencies of the original corpus.

We evaluate these controls in two complementary settings. First, in the two-stage diagnostic setting used throughout most of \Cref{res:when-and-why}, we construct a separate shuffled-token corpus for each of the five procedural tasks. Each model is pretrained on the shuffled corpus and then fine-tuned directly on Lipophilicity, without molecular pretraining. As shown in Appendix~\ref{app:shuffled-tokens}, every shuffled-token model loses the downstream benefit of its structured counterpart and performs no better than the randomly initialized fine-tuning baseline.

Second, we test whether this conclusion survives the full three-stage pipeline using \textsc{Dyck-Shuffle} as a representative control. The token-shuffled model undergoes the same molecular pretraining and downstream fine-tuning as the structured model. In \Cref{fig:molecular-pretraining-loss}, its held-out MTM loss remains worse than that of molecular pretraining from random initialization throughout all 12 epochs. After Lipophilicity fine-tuning, it obtains an MAE of \(0.4181 \pm 0.0062\), a roughly $2.46\%$ degradation relative to the molecular-pretraining-only baseline, whereas structured \textsc{Dyck-Shuffle} improves performance by $3.53\%$.


These results rule out generic exposure to synthetic sequences or their token-level statistics as a sufficient explanation for transfer. The downstream benefit depends on the structured relationships present in the procedural data.

\FloatBarrier
\subsection{\label{res:localisation}Where Does the Transfer Reside?}
The preceding experiments show that procedural structure 
can transfer to molecular prediction, but 
they do not identify where that information is represented within the transformer. We therefore ask which model components carry the transferable information learned during procedural pretraining.

We compare three transfer settings on Lipophilicity: transferring the full transformer backbone, transferring only the attention-layer weights, and transferring only the feed-forward-layer weights. All other parameters are reinitialized as described in \Cref{sec:methodology}.

\input{lipophilicity_full_backbone}
\Cref{fig:selective-transfer} shows that attention-only transfer matches or exceeds full-backbone transfer for \textsc{Dyck}, \textsc{Dyck-Shuffle}, and \textsc{Reverse}. 
In contrast, transferring only the feed-forward layers removes most of the downstream benefit and can even hurt performance.  \textsc{SSP} behaves differently: neither component alone recovers the full-transfer benefit, suggesting that useful information is distributed across both attention and feed-forward layers. \textsc{NCA} provides no significant benefit under any transfer configuration.

\input{layer_analysis}

The deterioration observed at large procedural budgets in \Cref{res:overtraining} raises a related question: does this over-specialization also reside in particular model components? To test this, we repeat the procedural-budget sweep while transferring only the attention-layer weights to the downstream task.

\Cref{fig:attention-robustness} shows that attention-only transfer is generally more robust to increasing procedural budgets than full-backbone transfer. For \textsc{Dyck} and \textsc{Dyck-Shuffle}, the downstream benefit no longer decays at the largest tested budgets when only attention weights are transferred. \textsc{Reverse} and \textsc{SSP} also decay more slowly under attention-only transfer, although both eventually converge back towards the baseline.

Together, these results indicate that much of the useful transferable structure learned from \textsc{Dyck}, \textsc{Dyck-Shuffle}, and \textsc{Reverse} is carried by the attention layers. They also suggest that part of the degradation caused by excessive procedural pretraining is associated with the feed-forward layers: discarding these weights preserves useful transfer over a wider range of procedural budgets. The separation is not complete, however, since \textsc{Reverse} and \textsc{SSP} still lose their advantage at sufficiently large budgets even under attention-only transfer.

%% file: merged-downstream-table.tex
\begin{table}[t]
\centering
\caption{Molecular property prediction after procedural and molecular pretraining:
All models are pretrained on our 250k molecule PubChem subset before downstream fine-tuning.
Standardized test MAE is reported (lower is better), with $\Delta$ relative to the molecular-pretraining-only baseline.
Results average $n{=}3$ seeds ($n{=}6$ for FreeSolv); best procedural results are bold.}
\label{tab:main_results}
\footnotesize
\setlength{\tabcolsep}{3.5pt}

\begin{tabular}{@{}l cc cc cc@{}}
\toprule
& \multicolumn{2}{c}{\textsc{Lipophilicity}}
& \multicolumn{2}{c}{\textsc{QM9 gap} (1k)}
& \multicolumn{2}{c}{\textsc{FreeSolv}} \\
\cmidrule(lr){2-3}
\cmidrule(lr){4-5}
\cmidrule(lr){6-7}

& MAE & $\Delta$\%
& MAE & $\Delta$\%
& MAE & $\Delta$\% \\
\midrule

Random init. $\rightarrow$ \textsc{PubChem250k} (baseline)
& 0.4083 {\scriptsize$\pm$0.0125} & ---
& 0.2662 {\scriptsize$\pm$0.0015} & ---
& 0.1676 {\scriptsize$\pm$0.0145} & --- \\

\addlinespace[3pt]
\multicolumn{7}{@{}l}{\textit{Procedural pretraining $\rightarrow$ \textsc{PubChem250k}}} \\

\quad \textsc{Dyck}
& 0.4028 {\scriptsize$\pm$0.0020} & $-1.29$
& 0.2641 {\scriptsize$\pm$0.0020} & $-0.80$
& 0.1764 {\scriptsize$\pm$0.0163} & $+5.69$ \\

\quad \textsc{Dyck-Shuffle}
& 0.3937 {\scriptsize$\pm$0.0029} & $-3.53$
& 0.2680 {\scriptsize$\pm$0.0026} & $+0.65$
& 0.1867 {\scriptsize$\pm$0.0129} & $+12.36$ \\

\quad \textsc{Reverse}
& \textbf{0.3885} {\scriptsize$\pm$0.0111} & $\mathbf{-4.80}$
& \textbf{0.2611} {\scriptsize$\pm$0.0004} & $\mathbf{-1.91}$
& \textbf{0.1621} {\scriptsize$\pm$0.0031} & $\mathbf{-2.59}$ \\

\quad \textsc{NCA}
& 0.4069 {\scriptsize$\pm$0.0027} & $-0.28$
& 0.2703 {\scriptsize$\pm$0.0054} & $+1.51$
& 0.1849 {\scriptsize$\pm$0.0110} & $+11.16$ \\

\quad \textsc{SMILES Shortest-Paths}
& 0.4011 {\scriptsize$\pm$0.0016} & $-1.70$
& 0.2671 {\scriptsize$\pm$0.0001} & $+0.32$
& 0.1713 {\scriptsize$\pm$0.0074} & $+3.13$ \\

\bottomrule
\end{tabular}
\end{table}

%% file: pretraining_dynamics_scarcity.tex
\begin{figure}[t]
  \centering
  \begin{subfigure}[t]{0.48\linewidth}
    \centering
    \includegraphics[width=\linewidth]{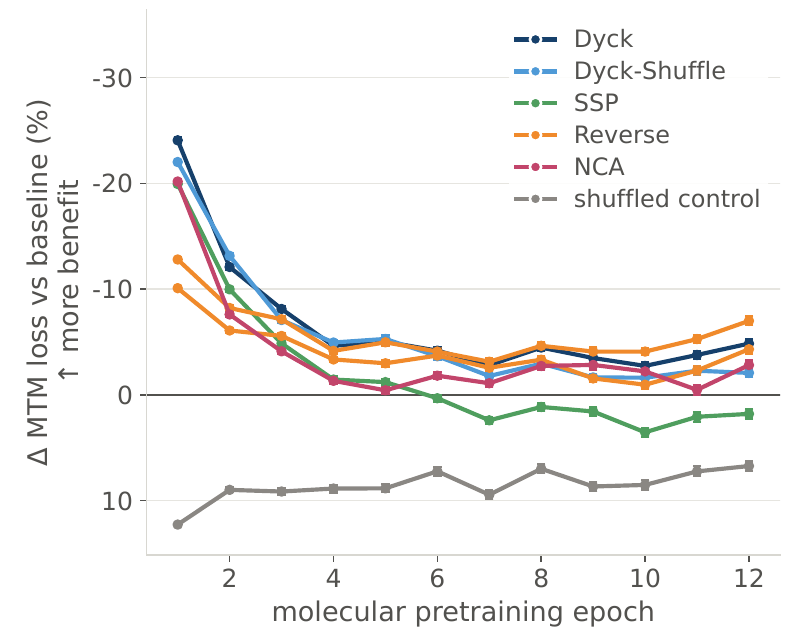}
    \caption{Held-out PubChem250k loss.}
    \label{fig:molecular-pretraining-loss}
  \end{subfigure}\hfill
  \begin{subfigure}[t]{0.48\linewidth}
    \centering
    \includegraphics[width=\linewidth]{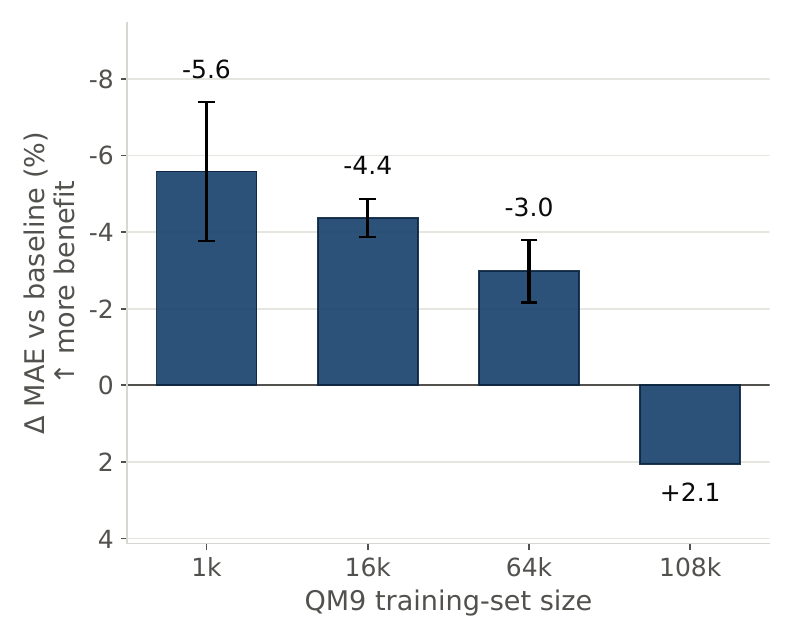}
    \caption{QM9 gap MAE across training-set sizes.}
    \label{fig:data-scarcity}
  \end{subfigure}
\caption{
Molecular pretraining dynamics and data-scarcity evaluation:
a) Held-out molecular pretraining loss for each procedural task at its best configuration; also includes a token-shuffled control model discussed in \Cref{exp:requires-structure}.
b) QM9 gap MAE across training-set sizes after \textsc{Dyck} ($k=64$) pretraining.
Values are relative to the randomly initialized baseline; error bars are $\pm1$ standard deviation over $n{=}3$ seeds. The full-dataset point in b) is a single seed. Panel (a) also includes the token-shuffled \textsc{Dyck-Shuffle} control discussed in \Cref{exp:requires-structure}}
  \label{fig:panels}
\end{figure}

%% file: downstream_pp_budgets.tex
\begin{figure}
  \centering
  \includegraphics[width=1\linewidth]{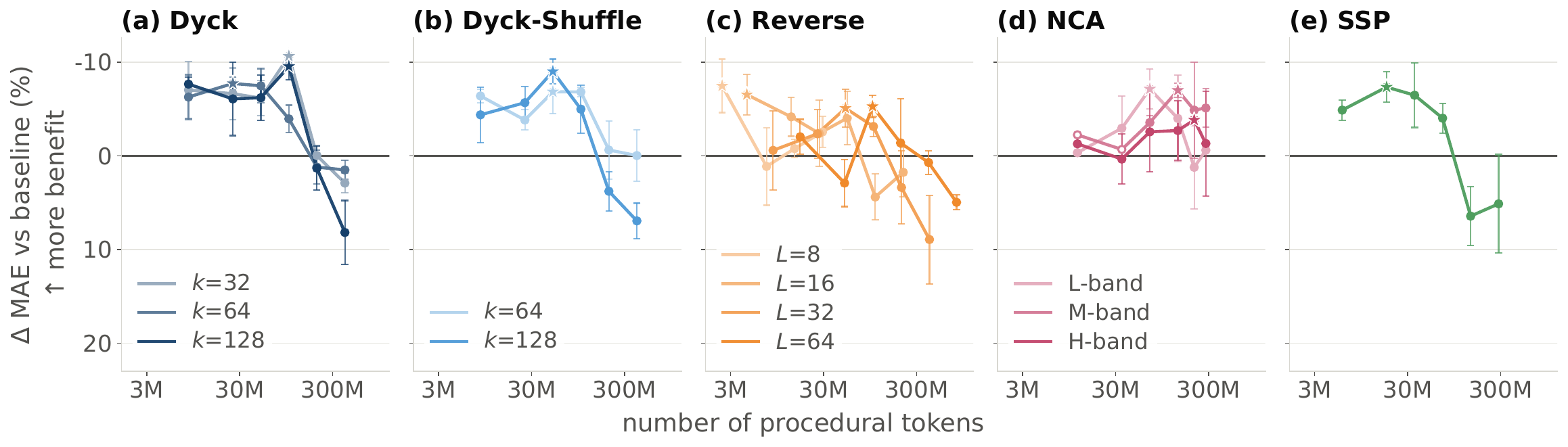}
  \caption{
  Downstream transfer across procedural pretraining budgets:
Lipophilicity performance relative to random initialization across procedural budgets and task-complexity settings. Error bars show $\pm1$ standard deviation over $n{=}3$ seeds; Across four of five tasks, downstream benefit vanishes at higher budgets. In particular, a few seeds at the highest budget collapsed to the loss of a constant predictor, these seeds are excluded from the mean calculation.}
  \label{fig:overtraining}
\end{figure}

%% file: lipophilicity_full_backbone.tex
\begin{wrapfigure}{r}{0.52\columnwidth}
    \centering
    \vspace{-8pt}
    \includegraphics[width=\linewidth]{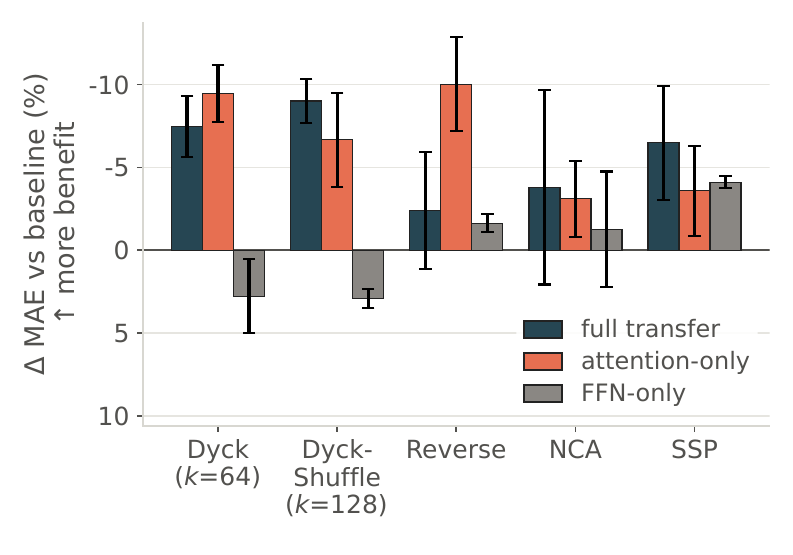}
    \caption{Lipophilicity benefit for full-backbone, attention-only, and feed-forward-only transfer. Attention-only transfer retains most of the benefit, while feed-forward-only transfer generally does not. }
    \label{fig:selective-transfer}
    \vspace{-8pt}
\end{wrapfigure}

%% file: layer_analysis.tex
\begin{figure}[b]
    \centering
    \includegraphics[width=\linewidth]{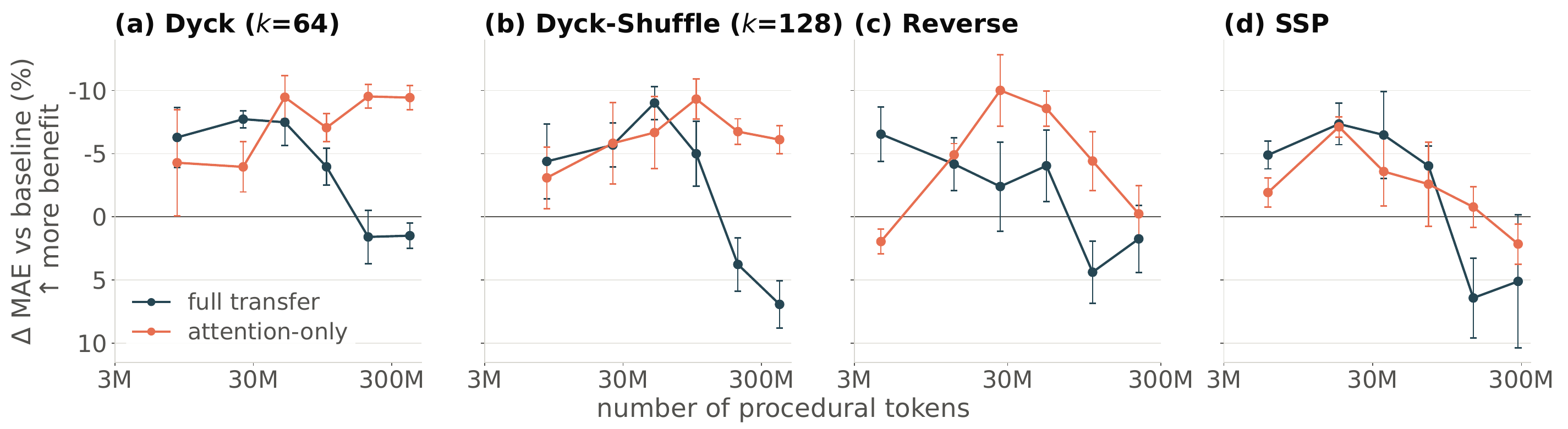}
    \caption{Full-backbone transfer compared with attention-only transfer at increasing procedural token budgets. We report Lipophilicity benefit over the randomly initialized baseline. Error bars are $\pm1$ standard deviation over $n=3$ seeds. Attention-only transfer decays more slowly than full transfer; \textsc{Dyck} and \textsc{Dyck-Shuffle} retain a benefit at the largest budgets, while \textsc{Reverse} and SSP still converge towards the baseline. }
    \label{fig:attention-robustness}
\end{figure}


%% file: discussion.tex
\section{Conclusion and Open Questions}
We showed that procedural pretraining on abstract, non-molecular data can improve molecular property prediction, even after subsequent pretraining on real molecular data. The gains are largest in low-data regimes, and in our strongest result, a simple sequence-reversal task recovers roughly 90\% of the performance difference between our 250K-molecule baseline and the publicly released MoLFormer checkpoint pretrained on approximately 100M molecules. The benefit depends on the structure of the procedural data rather than on simple token-level statistics: shuffling token order removes the transfer advantage.

Our results also show that more procedural pretraining is not always better. Across most tasks, downstream performance peaks at an intermediate training budget and deteriorates as the model approaches procedural convergence, suggesting a transition from learning broadly useful structure to task-specific specialization. For several tasks, much of the transferable information is retained in the attention layers, while the feed-forward layers appear to contribute more strongly to over-specialization.

Several questions remain open. What properties make a procedural task transferable, and can they be identified without downstream evaluation? Can the optimal stopping point for procedural pretraining be predicted from procedural training dynamics alone, rather than from downstream validation? Our results were also inconclusive on FreeSolv and for the NCA task, which showed no clear downstream benefit at any procedural budget. Determining whether these outcomes reflect dataset size, task predictability, or a more fundamental mismatch remains an important direction for future work. More broadly, do these effects persist across larger models, alternative molecular representations, and other scientific domains? Our findings suggest that improving molecular learning may require not only more molecular data, but also better ways of shaping the model before it sees a single molecule.

%% file: appendix.tex
\appendix

\section{Reproducibility and Training Details}
\subsection{Training details}

All three stages use the same MoLFormer encoder \citep{ross_large-scale_2022}: 12 layers, hidden size 768, 12 heads of dimension 64, feed-forward width 768, GELU activations, linear attention with 32 generalized random features, rotary position embeddings, a maximum length of 202 tokens and the released 2{,}362-token SMILES vocabulary (46.8M parameters with the MTM head, 48.0M with the regression head). All runs use one GPU (NVIDIA RTX 6000 24GB or A40 48GB), PyTorch 1.7.1 with PyTorch Lightning 1.1.5 and apex 22.03; procedural corpora are generated with numpy, RDKit and networkx (SSP) and the JAX generator of \citet{lee_training_2026} (NCA).

\Cref{tab:training-details} summarizes the training configuration for
each stage. All stages use \texttt{apex} FusedLAMB with
$\beta_1=0.9$, $\beta_2=0.99$, $\epsilon=10^{-6}$, no weight decay,
and global gradient-norm clipping at 1.0. The molecular-pretraining and fine-tuning configurations follow the released MoLFormer implementation~\citep{ross_large-scale_2022}, with three exceptions: for molecular pretraining we reduce the batch size from 1{,}200 to 256 and add a 500-step linear warm-up, and we fine-tune on QM9 for 400 rather than 500 epochs.

Token embeddings remain fixed at their random initialization during
procedural pretraining and are trainable during molecular pretraining
and downstream fine-tuning. Fine-tuning targets are standardized using
statistics computed from the training split, and the checkpoint with
the lowest validation MAE is used for test evaluation. Each procedural and each molecular pretraining is a single run with seed~12345. Fine-tuning uses seeds 12345, 42 and 43 (FreeSolv additionally uses 44, 45 and 46). The seed also draws the QM9 training subset.

\input{training-details-table}

\FloatBarrier
\subsection{Procedural-task generation}

\Cref{tab:procedural-details} summarizes the construction and masking
policy of each procedural task. The values evaluated for the task
complexity parameters are reported in \Cref{fig:validation-grid}. For \textsc{Reverse} and \textsc{SSP}, sequences are processed using the MoLFormer tokenizer and masked as in molecular pretraining: of the positions selected for prediction, 80\% are replaced by \texttt{[MASK]}, 10\% by a random token, and 10\% remain unchanged. \textsc{Dyck}, \textsc{Dyck-Shuffle}, and
\textsc{NCA} are represented directly as integer token IDs, and every
selected position is replaced by \texttt{[MASK]}.

For each task configuration and training budget of $B$ steps, we generate
$B\times256$ sequences once, using random seed 42, and train for a
single pass over them; the checkpoint at step $B$, the end of the cosine
schedule, is the one transferred, and no validation set is held out.

\input{procedural-details-table}

\FloatBarrier
\section{Choosing Task Complexity and Procedural Budget}

\begin{figure}[!h]
  \centering
  \includegraphics[width=0.88\linewidth]{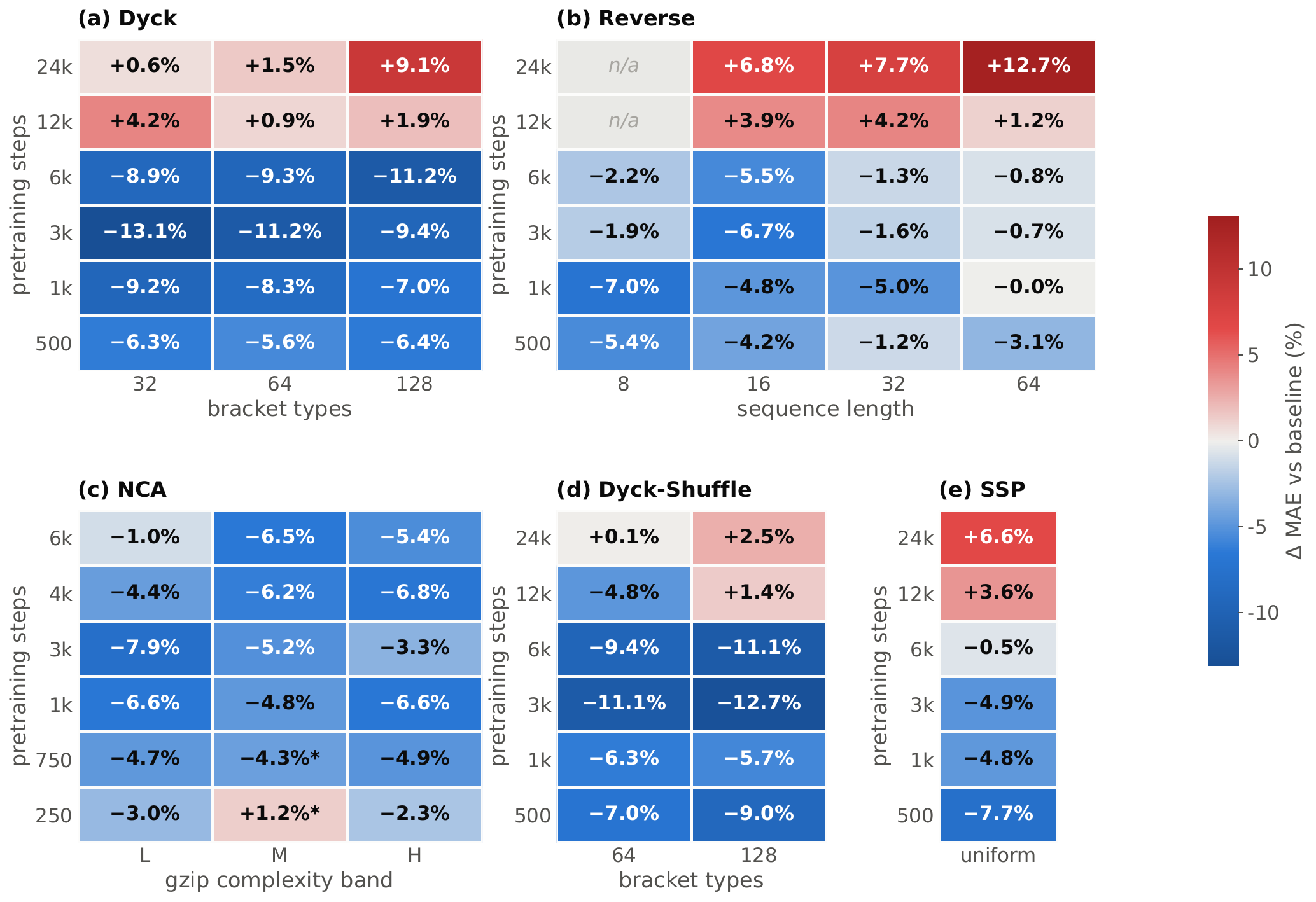}
  \caption{Grid search over task complexity and procedural-pretraining
  budget. Cells report mean validation-set Lipophilicity benefit over
  the randomly initialized baseline across \(n=3\) seeds.}
  \label{fig:validation-grid}
\end{figure}

\Cref{fig:validation-grid} shows the validation grid used to select
task complexity and procedural-pretraining budget for each task.

\FloatBarrier
\section{\label{app:bayes-floor}Procedural Pretraining Losses and Bayes-Floor Estimates}

In \Cref{tab:duration-sweep}, we report the procedural loss reached at every budget and the downstream benefit of the respective model on Lipophilicity over the random initialization baseline. We also mention the Bayes-optimal achievable loss for each task setting where we are able to determine it.

\input{procedural-loss-table}

\FloatBarrier
\clearpage
\section{\label{app:shuffled-tokens}Shuffling Token Order Removes Downstream Benefit Across All Procedural Tasks}
\begin{figure}[!ht]
    \centering
    \includegraphics[width=\linewidth]{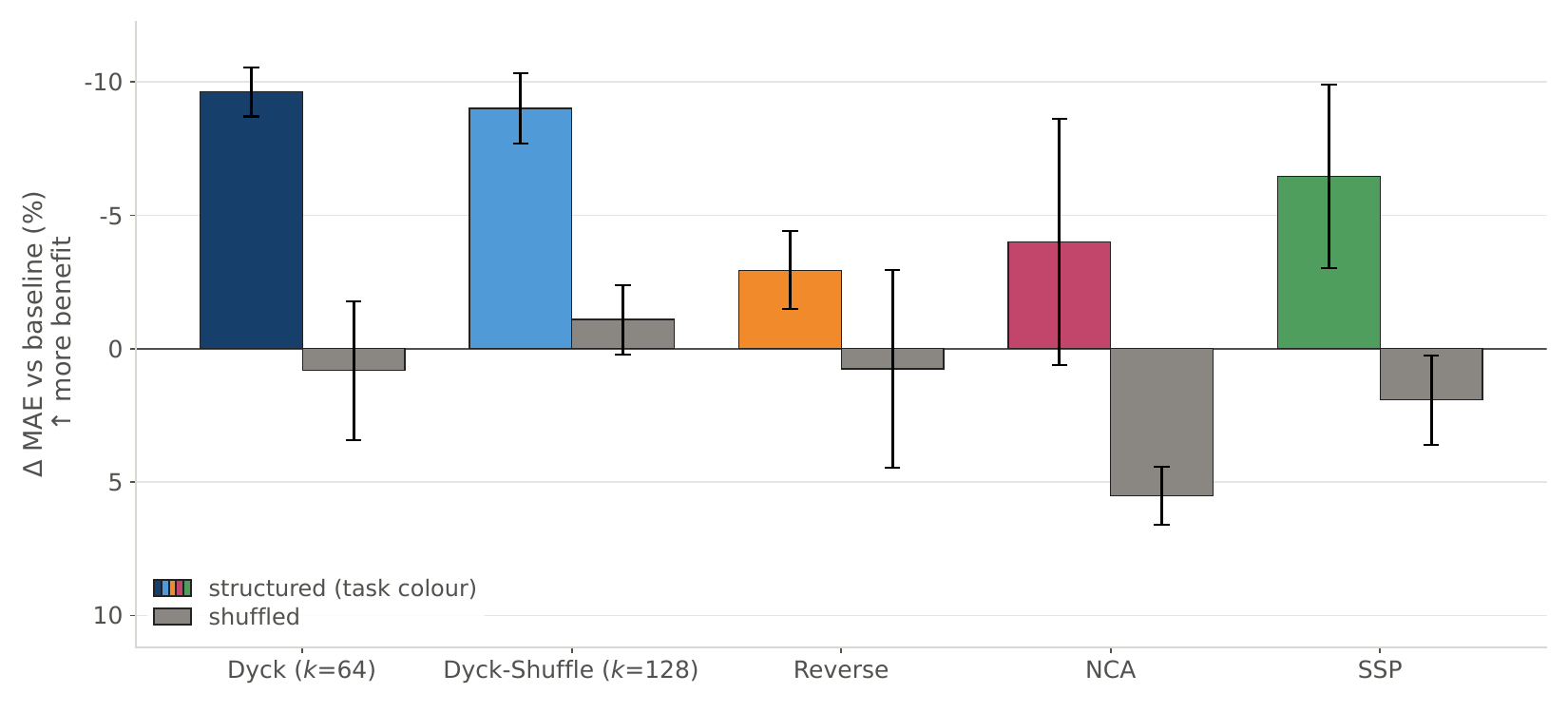}
    \caption{Token shuffling removes the downstream benefit of procedural pretraining across all five tasks in the two-stage diagnostic setting. Models are pretrained on either the structured corpus or a version in which tokens are shuffled within each sequence, and are then fine-tuned directly on Lipophilicity without molecular pretraining. Shuffling preserves each sequence's length and token counts while disrupting order-dependent structure, indicating that marginal token statistics alone are insufficient to explain the observed benefit.}
    \label{fig:shuffle-control}
\end{figure}

In \Cref{exp:requires-structure}, we found that shuffling the token order in the procedural sequences eliminates all downstream benefit. This section reports the shuffled-token ablation in the two-stage diagnostic setting for all five procedural tasks: procedural pretraining is directly followed by Lipophilicity fine-tuning. \Cref{fig:shuffle-control} shows that every procedural task loses its downstream benefit when the token order within the sequences is shuffled.

%% file: training-details-table.tex
\begin{table}[!ht]
\centering
\small
\setlength{\tabcolsep}{5pt}

\begin{tabularx}{\linewidth}{
    @{}l
    >{\raggedright\arraybackslash}X
    >{\raggedright\arraybackslash}X
    >{\raggedright\arraybackslash}X
    @{}
}
\toprule
Parameter & Procedural pretraining & Molecular pretraining & Downstream fine-tuning \\
\midrule
Objective
& MTM with task-specific masking
& MTM with 15\% of tokens selected for prediction
& Regression from the mean-pooled encoder output; $\ell_1$ loss on standardized targets \\

Peak learning rate
& $2.4\times10^{-4}$
& $2.4\times10^{-4}$
& $3\times10^{-5}$ \\

Learning-rate schedule
& 10\% linear warm-up, followed by cosine decay to 10\% of the peak rate
& 500-step linear warm-up, followed by a constant rate
& Constant; no warm-up \\

Batch size
& 256 sequences
& 256 molecules
& 128 molecules \\

Dropout probability
& 0.2
& 0.2
& 0.1\\

Training budget
& 500--24{,}000 steps
& 12 epochs
& 500 epochs for Lipophilicity and FreeSolv; 400 for QM9 \\
\bottomrule
\end{tabularx}
\caption{Training configurations for the three stages of the experimental pipeline.}
\label{tab:training-details}
\end{table}

%% file: procedural-details-table.tex
\begin{table}[!ht]
\centering
\small
\setlength{\tabcolsep}{4pt}
\begin{tabularx}{\linewidth}{
    @{}l
    >{\raggedright\arraybackslash}X
    >{\raggedright\arraybackslash}p{3.0cm}
    >{\raggedright\arraybackslash}p{2.8cm}
    @{}
}
\toprule
Task
& Generated sequence
& Complexity parameter
& Masking strategy \\
\midrule

\textsc{Reverse}
& A sequence of $L$ symbols drawn from an alphabet of 13 SMILES
atom tokens, followed by \texttt{.} and the same symbols in reverse
order
& Sequence length $L$
& Reversed tokens ($p_{\mathrm{mask}}=0.5$) \\

\textsc{Dyck}
& A 64-token balanced bracket string with $k$ distinct bracket types. An
opening bracket is sampled with probability 0.5 whenever permitted to form a valid balanced-bracket string
& Number of distinct bracket types $k$
& Closing brackets ($p_{\mathrm{mask}}=0.5$) \\

\textsc{Dyck-Shuffle}
& Similar to \textsc{Dyck}, except that a closing bracket may match any
currently open bracket of the same type, allowing bracket pairs to
cross
& Number of distinct bracket types $k$
& Closing brackets ($p_{\mathrm{mask}}=0.5$) \\

\textsc{SSP}
& The canonical SMILES representation of a random connected graph with
8--25 nodes and node degrees 1--4, followed by . and, for every node in SMILES order, its shortest-path distance to a source node marked as \texttt{*}
& --
& Distance tokens ($p_{\mathrm{mask}}=0.5$) \\

\textsc{NCA}
& Ten frames of an $8\times8$ cellular automaton with six cell states,
sampled two update steps apart. Each frame is represented by 16 tokens
corresponding to non-overlapping $2\times2$ patches
& gzip-compressibility band of the update rule
& Uniformly across all patch tokens ($p_{\mathrm{mask}}=0.15$) \\

\bottomrule
\end{tabularx}
\caption{Procedural-task generation and masking. For each task, we
report the generated sequence, the complexity parameter varied in our
experiments, and the tokens eligible for masking. Each eligible token
is selected independently with probability $p_{\mathrm{mask}}$.}
\label{tab:procedural-details}
\end{table}

%% file: procedural-loss-table.tex
\begin{table*}[!ht]\centering\footnotesize\setlength{\tabcolsep}{5pt}
\begin{tabular}{@{}llrrrrrrr@{}}\toprule
& & \multicolumn{6}{c}{Pretraining budget} & \\
\cmidrule(lr){3-8}
Task & Metric & $b_1$ & $b_2$ & $b_3$ & $b_4$ & $b_5$ & $b_6$ & $\mathcal{L}^\star$\\
\midrule
\multicolumn{9}{@{}l}{\textbf{Dyck}}\\
\quad $k{=}32$  & $\Delta$ MAE (\%)  & $-7.0$ & $-6.6$ & $-6.2$ & $-10.6$ & $-0.0$ & $+2.9$ & \\
                & procedural loss    & 1.46 & 0.72 & 0.266 & 0.0512 & $6.76{\times}10^{-3}$ & $5.62{\times}10^{-4}$ & $0$\\
\addlinespace[2pt]
\quad $k{=}64$  & $\Delta$ MAE (\%)  & $-6.3$ & $-7.7$ & $-7.5$ & $-4.0$ & $+1.2$ & $+1.5$ & \\
                & procedural loss    & 1.98 & 1.21 & 0.527 & 0.158 & 0.0144 & $8.73{\times}10^{-4}$ & $0$\\
\addlinespace[2pt]
\quad $k{=}128$ & $\Delta$ MAE (\%)  & $-7.7$ & $-6.1$ & $-6.2$ & $-9.6$ & $+1.3$ & $+13.3$ & \\
                & procedural loss    & 1.86 & 1.11 & 0.47 & 0.0943 & 1.45 & $6.81{\times}10^{-4}$ & $0$\\
\midrule
\multicolumn{9}{@{}l}{\textbf{Dyck-Shuffle}}\\
\quad $k{=}64$  & $\Delta$ MAE (\%)  & $-6.4$ & $-3.8$ & $-6.8$ & $-6.8$ & $-0.6$ & $-0.0$ & \\
                & procedural loss    & 2.12 & 1.58 & 1.44 & 1.34 & 1.27 & 1.26 & $1.250$\\
\addlinespace[2pt]
\quad $k{=}128$ & $\Delta$ MAE (\%)  & $-4.4$ & $-5.7$ & $-9.0$ & $-5.0$ & $+3.8$ & $+6.9$ & \\
                & procedural loss    & 2.36 & 1.82 & 1.57 & 1.41 & 1.31 & 1.28 & $1.267$\\
\midrule
\multicolumn{9}{@{}l}{\textbf{Reverse}}\\
\quad $L{=}8$   & $\Delta$ MAE (\%)  & $-7.5$ & $+1.1$ & $-0.8$ & $-2.6$ & -- & -- & \\
                & procedural loss    & 0.907 & $6.65{\times}10^{-4}$ & $4.39{\times}10^{-6}$ & $1.16{\times}10^{-5}$ & -- & -- & $0$\\
\addlinespace[2pt]
\quad $L{=}16$  & $\Delta$ MAE (\%)  & $-6.5$ & $-4.2$ & $-2.4$ & $-4.0$ & $+4.4$ & $+1.7$ & \\
                & procedural loss    & 1.65 & 0.229 & $1.07{\times}10^{-4}$ & $1.37{\times}10^{-5}$ & $6.44{\times}10^{-7}$ & $1.28{\times}10^{-7}$ & $0$\\
\addlinespace[2pt]
\quad $L{=}32$  & $\Delta$ MAE (\%)  & $-0.6$ & $-2.4$ & $-5.1$ & $-3.1$ & $+3.4$ & $+8.9$ & \\
                & procedural loss    & 2.03 & 1.27 & 0.302 & $1.68{\times}10^{-4}$ & $4.77{\times}10^{-5}$ & $2{\times}10^{-6}$ & $0$\\
\addlinespace[2pt]
\quad $L{=}64$  & $\Delta$ MAE (\%)  & $-2.0$ & $+2.9$ & $-5.3$ & $-1.4$ & $+0.7$ & $+5.0$ & \\
                & procedural loss    & 2.27 & 2.07 & 1.73 & 1.25 & 0.567 & 0.0474 & $0$\\
\midrule
\multicolumn{9}{@{}l}{\textbf{SSP}}\\
\quad uni       & $\Delta$ MAE (\%)  & $-4.9$ & $-7.4$ & $-6.5$ & $-4.0$ & $+6.4$ & $+5.1$ & \\
                & procedural loss    & 0.894 & 0.691 & 0.592 & 0.463 & 0.176 & 0.0622 & $0$\\
\midrule
\multicolumn{9}{@{}l}{\textbf{NCA}}\\
\quad band L    & $\Delta$ MAE (\%)  & $-0.3$ & $-3.0$ & $-7.1$ & $-4.0$ & $+1.2$ & $-0.6$ & \\
                & procedural loss    & 2.47 & 2.28 & 2.13 & 1.86 & 1.78 & 1.73 & unk.\\
\addlinespace[2pt]
\quad band M    & $\Delta$ MAE (\%)  & $-2.2$ & $-0.7$ & $-3.6$ & $-7.0$ & $-4.9$ & $-5.1$ & \\
                & procedural loss    & 3.75 & 3.58 & 3.46 & 3.41 & 3.36 & 3.27 & unk.\\
\addlinespace[2pt]
\quad band H    & $\Delta$ MAE (\%)  & $-1.3$ & $+0.3$ & $-2.6$ & $-2.7$ & $-3.8$ & $-1.3$ & \\
                & procedural loss    & 5.52 & 5.34 & 5.26 & 5.22 & 5.19 & 5.19 & unk.\\
\bottomrule
\end{tabular}
\caption{Procedural loss and downstream benefit for each tested task-complexity-duration combination. \(L^\star\) denotes the Bayes-optimal loss, which is unknown in closed form for NCA. Budgets \(b_1\)–\(b_6\) are \(\{500,1500,3000,6000,12000,24000\}\) optimizer steps for the sequence tasks and \(\{250,750,1500,3000,4500,6000\}\) for NCA.}
\label{tab:duration-sweep}
\end{table*}

%% file: references.bib
@inproceedings{jiang_procedural_2026,
  title={Procedural Pretraining: Warming Up Language Models with Abstract Data},
  author={Jiang, Liangze and Shinnick, Zachary and van den Hengel, Anton and Saratchandran, Hemanth and Teney, Damien},
  booktitle={Proceedings of the International Conference on Machine Learning (ICML)},
  year={2026},
}

@inproceedings{shinnick_can_2026,
  title={Can You Learn to See Without Images? Procedural Warm-Up for Vision Transformers},
  author={Shinnick, Zachary and Jiang, Liangze and Saratchandran, Hemanth and Teney, Damien and van den Hengel, Anton},
  booktitle={Proceedings of the IEEE/CVF Conference on Computer Vision and Pattern Recognition (CVPR)},
  year={2026},
}

@misc{lee_training_2026,
	title = {Training {Language} {Models} via {Neural} {Cellular} {Automata}},
	url = {http://arxiv.org/abs/2603.10055},
	doi = {10.48550/arXiv.2603.10055},
	urldate = {2026-06-26},
	publisher = {arXiv},
	author = {Lee, Dan and Han, Seungwook and Kumar, Akarsh and Agrawal, Pulkit},
	month = mar,
	year = {2026},
	note = {arXiv:2603.10055 [cs.LG]},
}

@misc{ross_large-scale_2022,
	title = {Large-{Scale} {Chemical} {Language} {Representations} {Capture} {Molecular} {Structure} and {Properties}},
	url = {http://arxiv.org/abs/2106.09553},
	doi = {10.48550/arXiv.2106.09553},
	urldate = {2026-07-29},
	publisher = {arXiv},
	author = {Ross, Jerret and Belgodere, Brian and Chenthamarakshan, Vijil and Padhi, Inkit and Mroueh, Youssef and Das, Payel},
	month = dec,
	year = {2022},
	note = {arXiv:2106.09553 [cs.LG]},
}

@misc{wu_moleculenet_2018,
	title = {{MoleculeNet}: {A} {Benchmark} for {Molecular} {Machine} {Learning}},
	shorttitle = {{MoleculeNet}},
	url = {http://arxiv.org/abs/1703.00564},
	doi = {10.48550/arXiv.1703.00564},
	urldate = {2026-08-05},
	publisher = {arXiv},
	author = {Wu, Zhenqin and Ramsundar, Bharath and Feinberg, Evan N. and Gomes, Joseph and Geniesse, Caleb and Pappu, Aneesh S. and Leswing, Karl and Pande, Vijay},
	month = oct,
	year = {2018},
	note = {arXiv:1703.00564 [cs.LG]},
}

@misc{hu_between_2025,
	title = {Between {Circuits} and {Chomsky}: {Pre}-pretraining on {Formal} {Languages} {Imparts} {Linguistic} {Biases}},
	shorttitle = {Between {Circuits} and {Chomsky}},
	url = {http://arxiv.org/abs/2502.19249},
	doi = {10.48550/arXiv.2502.19249},
	urldate = {2026-08-14},
	publisher = {arXiv},
	author = {Hu, Michael Y. and Petty, Jackson and Shi, Chuan and Merrill, William and Linzen, Tal},
	month = may,
	year = {2025},
	note = {arXiv:2502.19249 [cs.CL]},
}

@article{irwin_zinc_2004,
	title = {{ZINC} - {A} {Free} {Database} of {Commercially} {Available} {Compounds} for {Virtual} {Screening}},
	volume = {45},
	issn = {1549-9596},
	url = {https://doi.org/10.1021/ci049714+},
	doi = {10.1021/ci049714+},
	number = {1},
	urldate = {2026-08-23},
	journal = {Journal of Chemical Information and Modeling},
	author = {Irwin, John J. and Shoichet, Brian K.},
	month = dec,
	year = {2004},
	pages = {177--182},
}

@article{kim_pubchem_2019,
	title = {{PubChem} 2019 update: improved access to chemical data},
	volume = {47},
	issn = {0305-1048},
	shorttitle = {{PubChem} 2019 update},
	url = {https://doi.org/10.1093/nar/gky1033},
	doi = {10.1093/nar/gky1033},
	number = {D1},
	urldate = {2026-08-23},
	journal = {Nucleic Acids Research},
	author = {Kim, Sunghwan and Chen, Jie and Cheng, Tiejun and Gindulyte, Asta and He, Jia and He, Siqian and Li, Qingliang and Shoemaker, Benjamin A and Thiessen, Paul A and Yu, Bo and Zaslavsky, Leonid and Zhang, Jian and Bolton, Evan E},
	month = jan,
	year = {2019},
	pages = {D1102--D1109},
}

@misc{chithrananda_chemberta_2020,
	title = {{ChemBERTa}: {Large}-{Scale} {Self}-{Supervised} {Pretraining} for {Molecular} {Property} {Prediction}},
	shorttitle = {{ChemBERTa}},
	url = {http://arxiv.org/abs/2010.09885},
	doi = {10.48550/arXiv.2010.09885},
	urldate = {2026-08-24},
	publisher = {arXiv},
	author = {Chithrananda, Seyone and Grand, Gabriel and Ramsundar, Bharath},
	month = oct,
	year = {2020},
	note = {arXiv:2010.09885 [cs.LG]},
}

@inproceedings{baradad_jurjo_learning_2021,
	title = {Learning to {See} by {Looking} at {Noise}},
	volume = {34},
	url = {https://proceedings.neurips.cc/paper/2021/hash/14f2ebeab937ca128186e7ba876faef9-Abstract.html},
	urldate = {2026-08-24},
	booktitle = {Advances in {Neural} {Information} {Processing} {Systems}},
	publisher = {Curran Associates, Inc.},
	author = {Baradad Jurjo, Manel and Wulff, Jonas and Wang, Tongzhou and Isola, Phillip and Torralba, Antonio},
	year = {2021},
	pages = {2556--2569},
}

@misc{kataoka_pre-training_2021,
	title = {Pre-training without {Natural} {Images}},
	url = {http://arxiv.org/abs/2101.08515},
	doi = {10.48550/arXiv.2101.08515},
	urldate = {2026-08-25},
	publisher = {arXiv},
	author = {Kataoka, Hirokatsu and Okayasu, Kazushige and Matsumoto, Asato and Yamagata, Eisuke and Yamada, Ryosuke and Inoue, Nakamasa and Nakamura, Akio and Satoh, Yutaka},
	month = jan,
	year = {2021},
	note = {arXiv:2101.08515 [cs.CV]},
}

@misc{nomura_listen_2026,
	title = {Listen and {Chant} {Before} {You} {Read}: {The} {Ladder} of {Beauty} in {LM} {Pre}-{Training}},
	shorttitle = {Listen and {Chant} {Before} {You} {Read}},
	url = {http://arxiv.org/abs/2604.21265},
	doi = {10.5281/zenodo.19702183},
	urldate = {2026-08-25},
	author = {Nomura, Yoshinori},
	month = apr,
	year = {2026},
	note = {arXiv:2604.21265 [cs.CL]},
}

@inproceedings{papadimitriou_learning_2020,
	address = {Online},
	title = {Learning {Music} {Helps} {You} {Read}: {Using} {Transfer} to {Study} {Linguistic} {Structure} in {Language} {Models}},
	shorttitle = {Learning {Music} {Helps} {You} {Read}},
	url = {https://aclanthology.org/2020.emnlp-main.554/},
	doi = {10.18653/v1/2020.emnlp-main.554},
	urldate = {2026-08-25},
	booktitle = {Proceedings of the 2020 {Conference} on {Empirical} {Methods} in {Natural} {Language} {Processing} ({EMNLP})},
	publisher = {Association for Computational Linguistics},
	author = {Papadimitriou, Isabel and Jurafsky, Dan},
	editor = {Webber, Bonnie and Cohn, Trevor and He, Yulan and Liu, Yang},
	month = nov,
	year = {2020},
	pages = {6829--6839},
}

@misc{chiang_transferability_2021,
	title = {On the {Transferability} of {Pre}-trained {Language} {Models}: {A} {Study} from {Artificial} {Datasets}},
	shorttitle = {On the {Transferability} of {Pre}-trained {Language} {Models}},
	url = {https://arxiv.org/abs/2109.03537v2},
	language = {en},
	urldate = {2026-08-25},
	journal = {arXiv.org},
	author = {Chiang, Cheng-Han and Lee, Hung-yi},
	month = sep,
	year = {2021},
}

@inproceedings{wu_insights_2022,
	title = {Insights into {Pre}-training via {Simpler} {Synthetic} {Tasks}},
	url = {https://openreview.net/forum?id=XiLasGufCM},
	language = {en},
	urldate = {2026-08-25},
	author = {Wu, Yuhuai and Li, Felix and Liang, Percy},
	month = oct,
	year = {2022},
}

@article{shinnick2025transformers,
  title={Transformers pretrained on procedural data contain modular structures for algorithmic reasoning},
  author={Shinnick, Zachary and Jiang, Liangze and Saratchandran, Hemanth and Hengel, Anton van den and Teney, Damien},
  journal={arXiv preprint arXiv:2505.22308},
  year={2025}
}

@inproceedings{papadimitriou2023injecting,
  title={Injecting structural hints: Using language models to study inductive biases in language learning},
  author={Papadimitriou, Isabel and Jurafsky, Dan},
  booktitle={Findings of the Association for Computational Linguistics: EMNLP 2023},
  year={2023}
}

@misc{zhang2025intelligence,
      title={Intelligence at the Edge of Chaos}, 
      author={Shiyang Zhang and Aakash Patel and Syed A Rizvi and Nianchen Liu and Sizhuang He and Amin Karbasi and Emanuele Zappala and David van Dijk},
      year={2025},
      eprint={2410.02536},
      archivePrefix={arXiv},
      primaryClass={cs.AI},
      url={https://arxiv.org/abs/2410.02536}, 
}
